\documentclass[runningheads]{llncs}
\usepackage{graphicx}
\usepackage{amsmath,amssymb} 
\usepackage{color}
\usepackage{multirow}

\begin{document}
\pagestyle{headings}
\mainmatter

\def\ACCV20SubNumber{256}  

\title{Localize to Classify and Classify to Localize: \\ Mutual Guidance in Object Detection} 
\titlerunning{MutualGuide}
%
\author{Heng ZHANG\inst{1,3}\orcidID{0000-0001-6093-1729} \and
Elisa FROMONT \inst{1,4}\orcidID{0000-0003-0133-3491} \and
Sébastien LEFEVRE\inst{2}\orcidID{0000-0002-2384-8202} \and
Bruno AVIGNON\inst{3}}
\authorrunning{H. ZHANG et al.}
%
\institute{Univ Rennes, IRISA, France \and
Univ Bretagne Sud, IRISA, France \and
ATERMES company, France \and
IUF, Inria, France}

\maketitle

\begin{abstract}
Most deep learning object detectors are based on the anchor mechanism and resort to the Intersection over Union (IoU) between predefined anchor boxes and ground truth boxes to evaluate the matching quality between anchors and objects. In this paper, we question this use of IoU and propose a new anchor matching criterion guided, during the training phase, by the optimization of both the localization and the classification tasks: the predictions related to one task are used to dynamically assign sample anchors and improve the model on the other task, and vice versa. Despite the simplicity of the proposed method, our experiments with different state-of-the-art deep learning architectures on PASCAL VOC and MS COCO datasets demonstrate the effectiveness and generality of our Mutual Guidance strategy.
\end{abstract}

\section{Introduction}
\label{sec:intro}

Supervised object detection is a popular task in computer vision that aims at localizing objects through bounding boxes and assigning each of them to a predefined class. Deep learning-based methods largely dominate this research field and most recent methods are based on the anchor mechanism \cite{FasterRCNN,RFCN,FPN,PANet,LibraRCNN,CascadeRCNN,SSD,YOLOv3,RetinaNet,FSSD,RFBNet,M2Det}. 
Anchors are predefined reference boxes of different sizes and aspect ratios uniformly stacked over the whole image. They help the network to handle object scale and shape variations by converting the object detection problem into an anchor-wise bounding box regression and classification problem. Most state-of-the-art anchor-based object detectors resort to the Intersection over Union (IoU) between the predefined anchor boxes and the ground truth boxes (called $IoU_{anchor}$ in the following) to assign the sample anchors to an object (positive anchors) or a background (negative anchors) category. These assigned anchors are then used to minimize the bounding box regression and classification losses during training.

\begin{figure}[t]t
\begin{center}
\includegraphics[width=120mm]{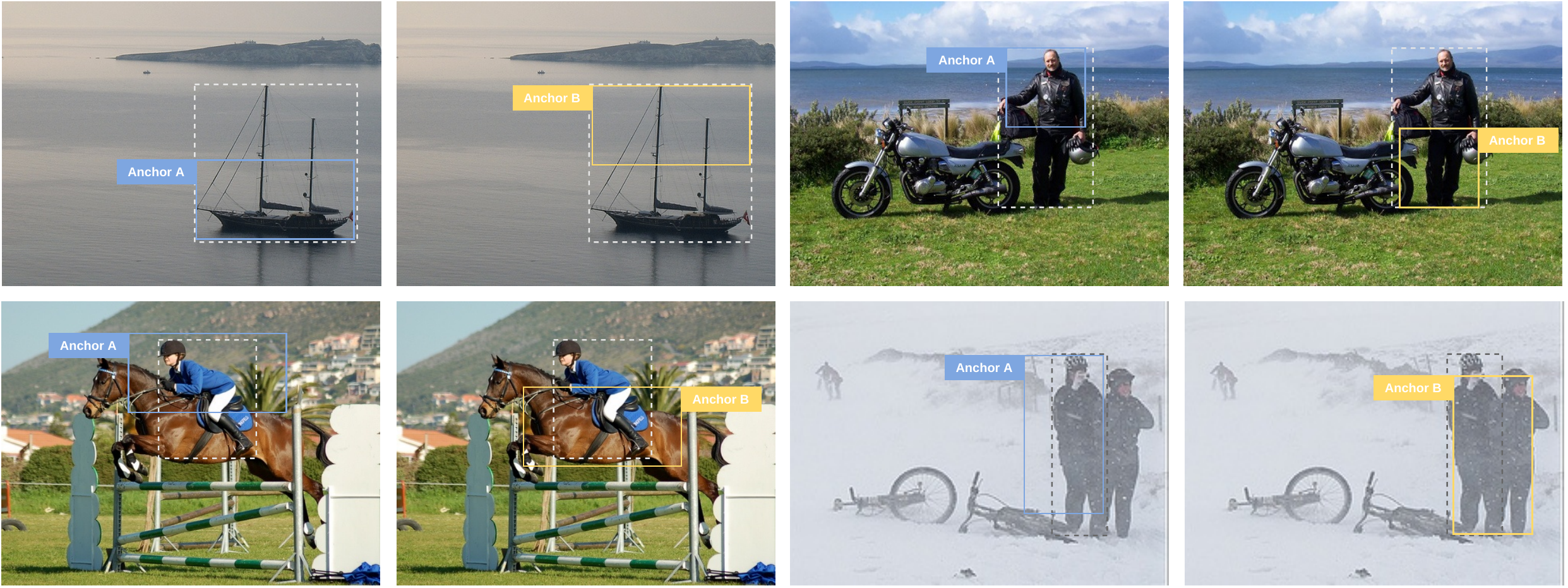}
\end{center}
\caption{Anchors A and anchors B have the same IoU with ground truth box but different visual semantic information. The ground truth in each image is marked as dotted-line box. Better viewed in colour.}
\label{fig:intro}
\end{figure}

This $IoU_{anchor}$-based anchor matching criterion is reasonable under the assumption that anchor boxes with high $IoU_{anchor}$ are appropriate for localization and classification. However, in reality, the $IoU_{anchor}$ is insensitive to objects' content/context, thus not ``optimal'' to be used, as such, for anchor matching.
In Figure \ref{fig:intro}, we show several examples where $IoU_{anchor}$ does not well reflect the matching quality between anchors and objects: anchors A and anchors B have exactly the same $IoU_{anchor}$ but possess very different matching qualities. For example, on the first line of Figure \ref{fig:intro}, anchors A covers a more representative and informative part of the object than anchors B; On the second line, anchors B contains parts of a nearby object which hinders the prediction on the jockey/left person.

Deep learning-based object detection involves two sub-tasks: instance localization and classification. Predictions for these two tasks tell us ``where'' and ``what'' objects are on the image respectively.
During the training phase, both tasks are jointly optimized by gradient descent, but the static anchor matching strategy does not explicitly benefit from the joint resolution of the two tasks, which may then yield to a task-misalignment problem, i.e., during the evaluation phase, the model might generate predictions with correct classification but imprecisely localized bounding boxes as well as predictions with precise localization but wrong classification. Both predictions significantly reduce the overall detection quality.

To address these two limitations of the existing $IoU_{anchor}$-based strategy, we propose a new, adaptive anchor matching criterion guided by the localization and by the classification tasks mutually, i.e., resorting to the bounding box regression prediction, we dynamically assign training anchor samples for optimizing classification and vice versa.
In particular, we constrain anchors that are well-localized to also be well-classified (\emph{Localize to Classify}), and those well-classified to also be well-localized (\emph{Classify to Localize}).
These strategies lead to a content/context-sensitive anchor matching and avoid the task-misalignment problem.
Despite the simplicity of the proposed strategy, \emph{Mutual Guidance} brings consistent Average Precision (AP) gains over the traditional static strategy with different deep learning architectures on PASCAL VOC \cite{VOC} and MS COCO \cite{COCO} datasets, especially on strict metrics such as AP75.
Our method is expected to be more efficient on applications that require a precise instance localization, e.g., autonomous driving, robotics, outdoor video surveillance, etc.

The rest of this paper is organized as follows: in Section \ref{sec:related}, we discuss some representative related work in object detection. Section \ref{sec:approach} provides implementation details of the proposed \emph{Mutual Guidance}. Section \ref{sec:expe} compares our dynamic anchor matching criterion to the traditional static criterion with different deep learning architectures on different public object detection datasets, and discusses reasons for the precision improvements. Section \ref{sec:conclusion} brings concluding remarks.

\section{Related work}
\label{sec:related}

Modern CNN-based object detection methods can be divided into two major categories: two-stage detectors and single-stage ones. Both categories give similar performance with a small edge in accuracy for the former and in efficiency for the latter.
Besides, both categories of detectors are massively based on the anchor mechanism which usually resorts to $IoU_{anchor}$ for evaluating the matching quality between  anchors and objects when assigning training labels and computing the bounding box regression and classification losses for a training example. 
Our method aims to improve this anchor matching criterion.

\subsection{Anchor-based object detection}

\paragraph{Two-stage object detectors.}
Faster RCNN \cite{FasterRCNN} defines the generic paradigm for two-stage object detectors: it first generates a sparse set of Regions of Interest (RoIs) with a Region Proposal Network (RPN), then classifies these regions and refines their bounding boxes. The RoIs are generated by the anchor mechanism.
Multiple improvements have been proposed based on this framework: 
R-FCN \cite{RFCN} suggests position-sensitive
score maps to share almost all computations on the entire image; 
FPN \cite{FPN} uses a top-down architecture and lateral connections to build high-level semantic feature maps at all scales; 
PANet \cite{PANet} enhances the multi-scale feature fusion by adding bottom-up path augmentation to introduce accurate localization signals in lower layers; 
Libra RCNN \cite{LibraRCNN} proposes the Balanced Feature Pyramid to further integrate multi-scale information into FPN; 
TridentNet \cite{TridentNet} constructs a parallel multi-branch architecture and adopts a scale-aware training scheme for training object scale specialized detection branches.
Cascade RCNN \cite{CascadeRCNN} further extends the two-stage paradigm into a multi-stage paradigm, where a sequence of detectors are trained stage by stage.

\paragraph{Single-stage object detectors.}
SSD \cite{SSD} and YOLO \cite{YOLOv1} are the fundamental methods for single-stage object detection. From this basis, many other works have been proposed:
FSSD \cite{FSSD} aggregates contextual information into the detector by concatenating features of different scales;
RetinaNet \cite{RetinaNet} proposes the Focal Loss to tackle the imbalanced classification problem that arises when trying to separate the actual object to detect from the massive background;
RFBNet \cite{RFBNet} proposes Receptive Field Block, which takes the relationship between the size and the eccentricity of the reception fields into account;
RefineDet \cite{RefineDet} introduces an additional stage of refinement for anchor boxes;
M2Det \cite{M2Det} stacks multiple thinned U-shape modules to tackle the so-called appearance-complexity variations.
While these methods introduce novel architectures to improve results for the object detection task, they all rely on the standard $IoU_{anchor}$-based matching. We identify this component as a possible limitation and propose a novel matching criterion, that could be adapted to any existing deep architecture for object detection.

\subsection{Anchor-free object detection}
The idea of anchor-free object detection consists in detecting objects not from predefined anchors boxes, but directly from particular key-points \cite{CornerNet,CornerNetLite,ExtremeNet,CenterNet2} or object centres \cite{CenterNet,FSAF,FCOS,FoveaBox}. 
However, these methods do not lead to a substantial accuracy advantage compared to anchor-based methods. The main idea of our \emph{Mutual Guidance} could also be applied to this class of object detectors, and the experimental results with anchor-free detectors are included in the supplementary material.

\section{Approach}
\label{sec:approach}

\begin{figure}[t]
\begin{center}
\includegraphics[width=120mm]{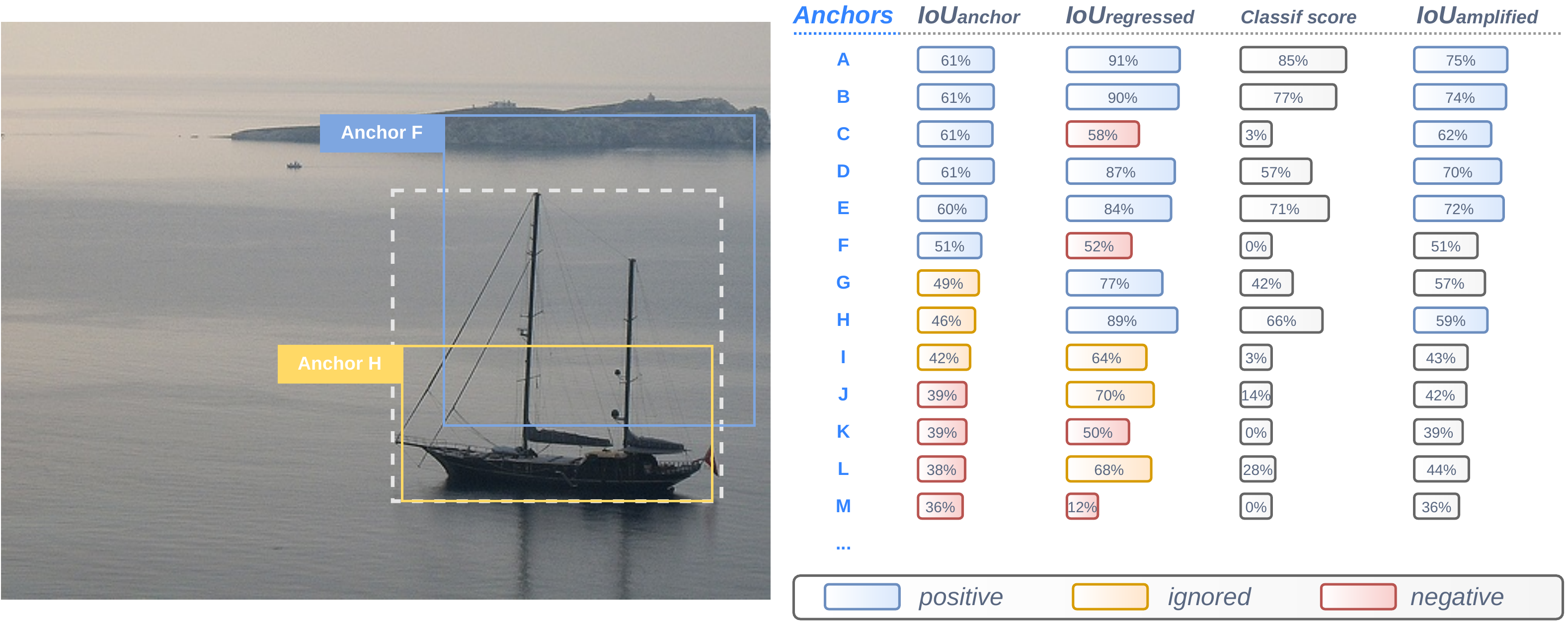}
\end{center}
   \caption{Illustration of different anchor matching strategies for the boat image resorting to $IoU_{anchor}$ (static), $IoU_{regressed}$ (\emph{Localize to Classify}) and $IoU_{amplified}$ (\emph{Classify to Localize}). Anchors A-M are predefined anchor boxes around the boat in the picture (only F and H are visualized as examples). Better viewed in colour.}
\label{fig:approach}
\end{figure}

As already sketched in the introduction, in order to train an anchor-based object detector, the predefined anchors should be assigned as \emph{positive} (``it is a true object'') or \emph{negative} (``it is a part of the background'') according to an evaluation of the matching between the anchors and the ground truth objects. Then, the bounding box regression loss is optimized according to the positive anchors, and the instance classification loss is optimized according to the positive as well as the negative anchors.
When training an anchor-based single-stage object detector with a static anchor matching strategy, the IoU between predefined anchor boxes and ground truth boxes ($IoU_{anchor}$) is the usual matching criterion.
As shown in the $IoU_{anchor}$ column of Figure \ref{fig:approach}, anchors with more than 50\% of $IoU_{anchor}$ are labelled as ``positive'', those with less than 40\% of $IoU_{anchor}$ are labelled as ``negative'', the rest are ``ignored anchors''. Note that at least one anchor should be assigned as positive, hence if there is no anchor with more than 50\% of $IoU_{anchor}$, the anchor with the highest $IoU_{anchor}$ is considered.

The proposed \emph{Mutual Guidance} consists of two components: \emph{Localize to Classify} and \emph{Classify to Localize}.

\subsection{Localize to Classify}
\label{sec:approach_r2c}

If an anchor is capable to precisely localize an object, this anchor must cover a good part of the semantically important area of this object and thus could be considered as an appropriate positive sample for classification. Drawing on this, we propose to leverage the IoU between regressed bounding boxes (i.e., the network's localization predictions) and ground truth boxes (noted $IoU_{regressed}$) to better assign the anchor labels for classification.
Inspired by the usual $IoU_{anchor}$, we compare $IoU_{regressed}$ to some given thresholds (discussed in the next paragraph) and then define anchors with $IoU_{regressed}$ greater than a high threshold as positive samples, and those with $IoU_{regressed}$ lower than a low threshold as negative samples (see $IoU_{regressed}$ column of Figure \ref{fig:approach}).

We now discuss a dynamic solution to set the thresholds. A fixed threshold (e.g., 50\% or 40\%) does not seem optimal since the network's localization ability gradually improves during the training procedure and so does the $IoU_{regressed}$ for each anchor, leading to the assignment of more and more positive anchors which destabilizes the training.
To address this issue, we propose a dynamic thresholding strategy. 
Even though the $IoU_{anchor}$ is not the best choice to accurately indicate the matching quality between anchors and objects, the number of assigned \emph{positive} and \emph{ignored} anchors does reflect the global matching conditions (brought by the size and the aspect ratio of the objects to detect), thus these numbers could be considered as reference values for our dynamic criterion.
As illustrated in Figure \ref{fig:approach}, while applying the $IoU_{anchor}$-based anchor matching strategy with the thresholds being 50\% and 40\%, the number of positive anchors ($N_p$) and ignored anchors ($N_i$) are noted ($N_p=6$ and $N_i=3$ for the boat). We then use these numbers to label the $N_p$ highest $IoU_{regressed}$ anchors as positive, and the following $N_i$ anchors as ignored. 
More formally, we exploit the $N_p$-th largest $IoU_{regressed}$ as our positive anchor threshold, and the ($N_p+N_i$)-th largest $IoU_{regressed}$ as our ignored anchor threshold.
Using this, our \emph{Localize to Classify} anchor matching strategy evolves with the network's localization capacity and maintains a consistent number of anchor samples assigned to both categories (positive/negative) during the whole training procedure.

\subsection{Classify to localize}
\label{sec:approach_c2r}

\begin{figure}[t]
\begin{center}
\includegraphics[width=120mm]{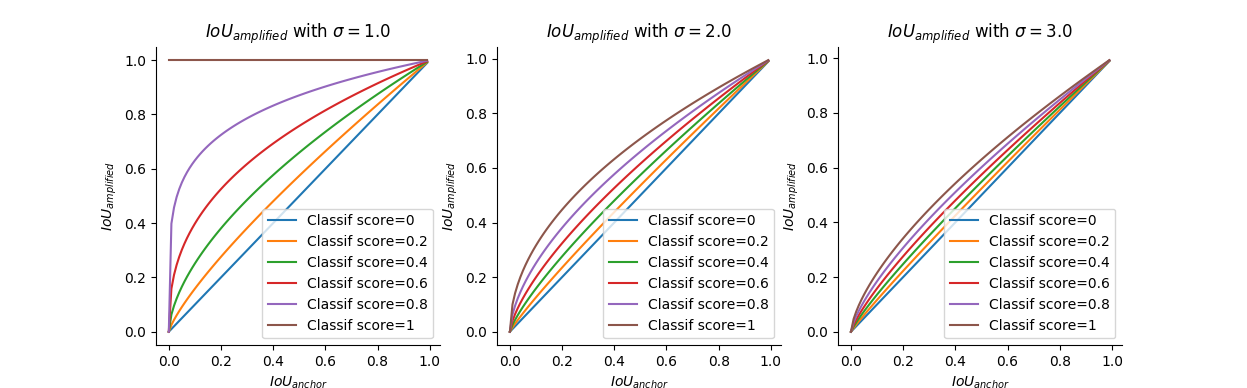}
\end{center}
   \caption{Illustration of $IoU_{amplified}$ with different $\sigma$ values (1, 2 or 3). $IoU_{amplified} = IoU_{anchor}$ when $Classif\ score = 0$.}
\label{fig:iouamp}
\end{figure}

As with the \emph{Localize to Classify} process, the positive anchor samples in \emph{Classify to Localize} are assigned according to the network's classification predictions (noted $Classif\ score$). 
Specifically, $Classif\ score$ is the predicted classification score for the object category, e.g., the $Classif\ score$ of Figure \ref{fig:approach} indicates the classification score for the \emph{boat} category. 

Nevertheless, this $Classif\ score$ is not effective enough to be used directly for assigning good positive anchors for the bounding box regression optimization. It is especially true at the beginning of the training process, when the network's weights are almost random values and all predicted classification scores are close to zero. 
The $IoU_{regressed}$ is optimized on the basis of the $IoU_{anchor}$, therefore we have $IoU_{regressed} \geq IoU_{anchor}$ in most cases (even at the beginning of the training), and this property helps to avoid such cold start problem and ensures training stability.
Symmetrically to the \emph{Localize to Classify} strategy, we now propose a \emph{Classify to Localize} strategy based on an $IoU_{amplified}$ defined as:
\begin{equation}\label{eq:1}
    IoU_{amplified} = (IoU_{anchor})^{\frac{\sigma-p}{\sigma}}
\end{equation}
where $\sigma$ is a hyper-parameter aiming at adjusting the degree of amplification, $p$ represents the mentioned $Classif\ score$. Inspired by the “focal loss” \cite{RetinaNet}, we chose eq. \ref{eq:1} as the simplest one  able  to  amplify  the  IoU  of  anchors  according  to  the  correct  classification predictions $p$. Its behavior is shown in Figure \ref{fig:iouamp}. The $IoU_{amplified}$ is always higher than the $IoU_{anchor}$, and the amplification is proportional to the predicted $Classif\ score$. In particular, the amplification is stronger for smaller $\sigma$ (note that $\sigma$ should be larger than 1), and disappears when $\sigma$ becomes large.

Similarly to the \emph{Localize to Classify} strategy, we apply a dynamic thresholding strategy to keep the number of assigned positive samples for the localization task and for the classification task consistent, e.g., we assign in Figure \ref{fig:approach}, the top $6$ anchors with the highest $IoU_{amplified}$ as positive samples. Note that there is no need for selecting ignored or negative anchors for the localization task since the background does not have an associated ground truth box.

As discussed in Section \ref{sec:intro}, $IoU_{anchor}$ is not sensitive to the content or the context of an object. Our proposed \emph{Localize to Classify} and \emph{Classify to Localize}, however, attempt to adaptively label the anchor samples according to their visual content and context information.
Considering anchor F and anchor H in Figure \ref{fig:approach}, one can tell that anchor H is better than anchor F for recognizing this boat, even with a smaller $IoU_{anchor}$. Using both our strategies, anchor H has been promoted to positive thanks to its excellent prediction quality on both tasks whereas anchor F has been labelled as negative even though it has a large $IoU_{anchor}$.

\subsection{About the task-misalignment problem}
\label{sec:align}

Since \emph{Localize to Classify} and \emph{Classify to Localize} are independent strategies, they could possibly assign contradictory positive/negative labels (e.g, the anchor C in Figure \ref{fig:approach} is labelled negative for the classification task but positive for the bounding box regression task). This happens when one anchor entails a good prediction on one task and a poor prediction on the other (i.e. they are misaligned predictions). Dealing with such contradictory labels, as we do with \emph{Mutual Guidance}, does not harm the training process. On the contrary, our method tackles the task-misalignment problem since the labels for one task are assigned according to the prediction quality on the other task, and vice versa. This mechanism forces the network to generate aligned predictions: if the classification prediction from one anchor is good while its localization prediction is bad, the \emph{Mutual Guidance} will give a positive label on the localization task to this anchor, to constrain it to be better at localizing as well while giving a negative label (i.e. background) on the classification task to avoid misaligned predictions.
In fact, the predicted classification score of this mislocalized anchor should be low enough for the anchor to be suppressed by the NMS procedure in the inference phase. The same reasoning holds for a good localization prediction with a bad classification one.

On the contrary, if a network always assigns similar positive/negative labels (as done in standard $IoU_{anchor}$-based methods) to both tasks during training, one cannot guarantee that there will be no misalignment of the localization and the classification predictions at inference time. 
Keeping anchors (after NMS) with misaligned predictions is harmful for strict evaluation metrics such as AP75.
\section{Experiments}
\label{sec:expe}

\subsection{Experimental Setting}

\paragraph{Network architecture and parameters.}
In order to test the generalization performance of the proposed method, we implement our method on the single-stage object detectors FSSD \cite{FSSD}, RetinaNet \cite{RetinaNet} and RFBNet \cite{RFBNet} using both ResNet-18 \cite{ResNet} or VGG-16 \cite{VGG} as backbone networks in our experiments. Note that RFBNet is not implemented with ResNet-18 as backbone since the two architectures are not compatible.
The backbone networks are pre-trained on ImageNet-1k classification dataset \cite{ImageNet}. 
We adopt the Focal Loss \cite{RetinaNet} and Balanced L1 Loss \cite{LibraRCNN} as our instance classification and bounding box regression loss functions respectively for all experiments.
The input image resolution is fixed to $320\times320$ pixels for all experiments (single scale training and evaluation).
Unless specified, all other implementation details are the same as in \cite{RFBNet}.
Following the results of Figure \ref{fig:iouamp}, we decided to fix our only new hyper-parameter $\sigma$ to $2$ for all experiments. $\sigma$ is used to set the degree of amplification when computing $IoU_{amplified}$ in Eq.~\eqref{eq:1}. It needs to be greater than $1$ for the exponent to be positive and lower than $3$ since this does not bring any amplification as shown in Figure \ref{fig:iouamp}. 

\paragraph{Datasets and evaluation metrics.}
Extensive experiments are performed on two benchmark datasets: PASCAL VOC \cite{VOC} and MS COCO \cite{COCO}. PASCAL VOC dataset has 20 object categories. Similarly to previous works, we utilize the combination of VOC2007 and VOC2012 trainval sets for training, and rely on the VOC2007 test for evaluation. MS COCO dataset contains 80 classes. Our experiments on this dataset are conducted on the train2017 and val2017 set for training and evaluation respectively.
For all datasets, we use the evaluation metrics introduced in the MS COCO benchmark: the Average Precision (AP) averaged over 10 IoU thresholds from 0.5 to 0.95, but also AP50, AP75, AP$_s$, AP$_m$, AP$_l$. AP50 and AP75 measure the average precision for a given IoU threshold (50\% and 75\%, respectively). The last three aim at focusing on small ($area < 32^2$), medium ($32^2 < area < 96^2$) and large ($area > 96^2$) objects respectively. Since the size of the objects greatly varies between MS COCO and PASCAL VOC, these size-dependent measures are ignored when experimenting with PASCAL VOC dataset.


\subsection{Results}
\paragraph{Experiments on PASCAL VOC.}

\begin{table}[t]
\begin{center}
\begin{tabular}{l|c|ccc}
\hline
Model & Matching strategy & $AP$ & $AP50$ & $AP75$ \\ \hline\hline
\multirow{4}{*}{\begin{tabular}[c]{@{}l@{}}FSSD with \\ ResNet-18 backbone\end{tabular}} & $IoU_{anchor}$-based & 50.3\% & 75.5\% & 53.7\%  \\
 & \emph{Localize to Classify}  & 51.8\% & 76.1\% & 55.9\% \\
 & \emph{Classify to Localize}  & 51.0\% & 76.1\% & 54.3\% \\
 & \emph{Mutual Guidance} & \textbf{52.1\%} & \textbf{76.2\%} & \textbf{55.9\%} \\ \hline
\multirow{4}{*}{\begin{tabular}[c]{@{}l@{}}FSSD with \\ VGG-16 backbone\end{tabular}} & $IoU_{anchor}$-based & 54.1\% & 80.1\% & 58.3\% \\
 & \emph{Localize to Classify}  & 56.0\% & 80.3\% & 60.6\% \\
 & \emph{Classify to Localize}  & 54.4\% & 79.9\% & 58.5\% \\
 & \emph{Mutual Guidance} & \textbf{56.2\%} & \textbf{80.4\%} & \textbf{61.4\%} \\ \hline
 \multirow{4}{*}{\begin{tabular}[c]{@{}l@{}}RetinaNet with \\ ResNet-18 backbone\end{tabular}} & $IoU_{anchor}$-based & 51.1\% & 75.8\% & 54.8\%  \\
 & \emph{Localize to Classify}  & 53.4\% & 76.5\% & 57.2\% \\
 & \emph{Classify to Localize}  & 51.9\% & 75.9\% & 55.8\% \\
 & \emph{Mutual Guidance} & \textbf{53.5\%} & \textbf{76.9\%} & \textbf{57.4\%} \\ \hline
\multirow{4}{*}{\begin{tabular}[c]{@{}l@{}}RetinaNet with \\ VGG-16 backbone\end{tabular}} & $IoU_{anchor}$-based & 55.2\% & 80.2\% & 59.6\% \\
 & \emph{Localize to Classify}  & 57.4\% & 81.1\% & 62.6\% \\
 & \emph{Classify to Localize}  & 56.2\% & 80.1\% & 61.7\% \\
 & \emph{Mutual Guidance} & \textbf{57.7\%} & \textbf{81.1\%} & \textbf{62.9\%} \\ \hline
\multirow{4}{*}{\begin{tabular}[c]{@{}l@{}}RFBNet with \\ VGG-16 backbone\end{tabular}} & $IoU_{anchor}$-based & 55.6\% & 80.9\% & 59.6\% \\
 & \emph{Localize to Classify} & 57.2\% & 80.9\% & 61.6\% \\
 & \emph{Classify to Localize}  & 55.9\% & 80.8\% & 60.2\% \\
 & \emph{Mutual Guidance} & \textbf{57.9\%} & \textbf{81.5\%} & \textbf{62.6\%} \\ \hline
\end{tabular}
\end{center}
\caption{Comparison of different anchor matching strategies (the usual $IoU_{anchor}$-based, proposed \emph{Localize to Classify}, \emph{Classify to Localize} and \emph{Mutual Guidance}) for object detection. Experiments are conducted on the PASCAL VOC dataset. The best score for each architecture is in bold.}
\label{tab:voc}
\end{table}

We evaluate the effectiveness of both components (\emph{Localize to Classify} and \emph{Classify to Localize}) of our proposed approach w.r.t. the usual $IoU_{anchor}$-based matching strategy when applied on the same deep learning architectures. 
The results obtained on the PASCAL VOC dataset are given in Table \ref{tab:voc}.
Both proposed anchor matching strategies consistently boost the performance of the ``vanilla'' networks and their combination (\emph{Mutual Guidance}) leads to the best AP and all other evaluation metrics. 

In particular, we observe that the improvements are small on AP50 (around 0.5\%) but significant on AP75 (around 3\%), which means that we obtain \emph{more precise detections}. As analysed in Section \ref{sec:align}, this comes from the task-misalignment problem faced with the usual static anchor matching methods. This issue leads to retain well-classified but poorly-localized predictions and suppress well-localized but poorly-classified predictions, which in turns results in a significant drop of the AP score at strict IoU thresholds, e.g., AP75. 
In \emph{Mutual Guidance}, however, training labels for one task are dynamically assigned according to the prediction quality on the other task and vice versa. This connection makes the classification and localization tasks consistent along all training phases and as such avoids this task-misalignment problem.

We also notice that \emph{Localize to Classify} alone brings, for all five architectures, a higher improvement than \emph{Classify to Localize} alone. We hypothesize two possible reasons for this: 1) most object detection errors come from wrong classification instead of imprecise localization, so the classification task is more difficult than the localization task and thus, there is more room for the improvement on this task; 2) the amplification proposed in Eq.~\eqref{eq:1} may not be the most appropriate one to take advantage of the classification task for optimizing the bounding box regression task.

\paragraph{Experiments on MS COCO.}

\begin{table}
\begin{center}
\begin{tabular}{l|c|cccccc}
\hline
Model & Matching strategy & $AP$ & $AP50$ & $AP75$ & $AP_{s}$ & $AP_{m}$ & $AP_{l}$ \\ \hline\hline
\multirow{2}{*}{\begin{tabular}[c]{@{}l@{}}FSSD with \\ ResNet-18 backbone\end{tabular}} & $IoU_{anchor}$-based & 26.1\% & 42.8\% & 26.7\% & 8.6\% & 29.1\% & 41.0\% \\
 & \emph{Mutual Guidance} & \textbf{27.0\%} & \textbf{42.9\%} & \textbf{28.2\%} & \textbf{9.5\%} & \textbf{29.7\%} & \textbf{43.0\%} \\ \hline
\multirow{2}{*}{\begin{tabular}[c]{@{}l@{}}FSSD with \\ VGG-16 backbone\end{tabular}} & $IoU_{anchor}$-based & 31.1\% & 48.9\% & 32.7\% & 13.3\% & 37.2\% & 44.7\% \\
 & \emph{Mutual Guidance} &  \textbf{32.0\%} & \textbf{49.3\%} & \textbf{33.9\%} & \textbf{13.7\%} & \textbf{37.8\%} & \textbf{46.4\%} \\ \hline
\multirow{2}{*}{\begin{tabular}[c]{@{}l@{}}RetinaNet with \\ ResNet-18 backbone\end{tabular}} & $IoU_{anchor}$-based & 27.8\% & 44.5\% & 28.6\% & 10.4\% & 31.6\% & 42.6\% \\
 & \emph{Mutual Guidance} & \textbf{28.7\%} & \textbf{44.9\%} & \textbf{29.9\%} & \textbf{11.0\%} & \textbf{32.2\%} & \textbf{44.8\%} \\ \hline
\multirow{2}{*}{\begin{tabular}[c]{@{}l@{}}RetinaNet with \\ VGG-16 backbone\end{tabular}} & $IoU_{anchor}$-based & 32.3\% & 50.3\% & 34.0\% & 14.3\% & 37.9\% & 46.7\% \\
 & \emph{Mutual Guidance} &  \textbf{33.6\%} & \textbf{50.8\%} & \textbf{35.7\%} & \textbf{15.4\%} & \textbf{38.9\%} & \textbf{48.8\%} \\ \hline
\multirow{2}{*}{\begin{tabular}[c]{@{}l@{}}RFBNet with \\ VGG-16 backbone\end{tabular}} & $IoU_{anchor}$-based &  33.4\% & 51.6\% & 35.1\% & 14.2\% & 38.3\% & 49.1\% \\
 & \emph{Mutual Guidance} &  \textbf{34.6\%} & \textbf{52.0\%} & \textbf{36.8\%} & \textbf{15.8\%} & \textbf{39.0\%} & \textbf{51.1\%} \\ \hline
\end{tabular}
\end{center}
\caption{AP performance of different architectures for object detection on MS COCO dataset using 2 different anchor matching strategies: the usual $IoU_{anchor}$-based one and our complete approach marked as \emph{Mutual Guidance}. The best score for each architecture is in bold.}
\label{tab:sota}
\end{table}

We then conduct experiments on the more difficult MS COCO \cite{COCO} dataset and report our results in Table \ref{tab:sota}.
Note that according to the scale range defined by MS COCO, APs of small, medium and large objects are listed. In this dataset also, our \emph{Mutual Guidance} strategy consistently brings some performance gains compared to the $IoU_{anchor}$-based baselines. We notice that our AP gains on large objects is significant (around 2\%). This is because larger objects generally have more matched positive anchors, which offers more room for improvements to our method. Since the \emph{Mutual guidance} strategy only involves the training phase, and since there is no difference between $IoU_{anchor}$-based and our method during the evaluation phase, these improvements can be considered cost-free.

\subsection{Qualitative analysis}

\paragraph{Label assignment visualization.}

\begin{figure}[t]
\begin{center}
\includegraphics[width=120mm]{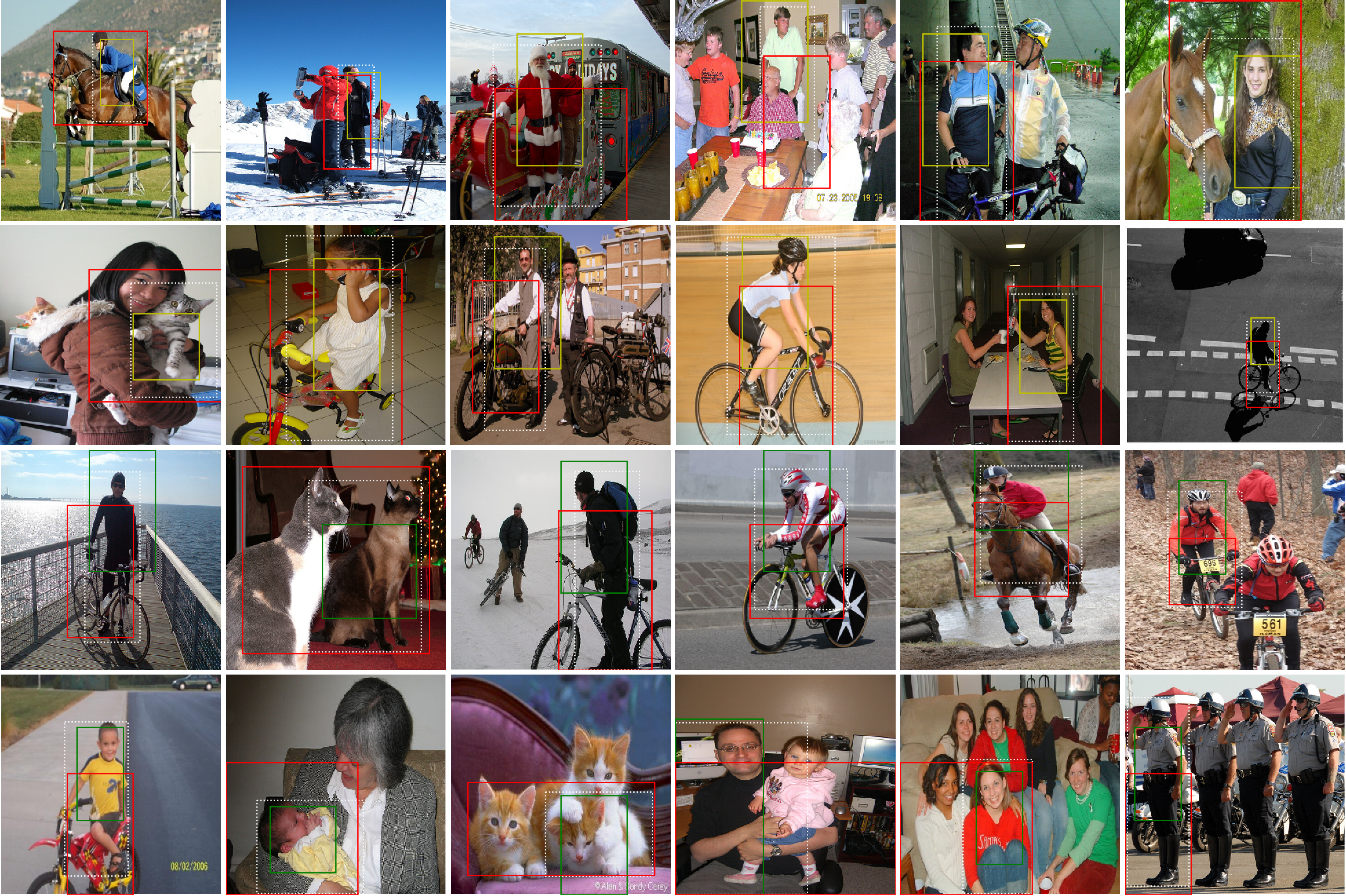}
\end{center}
   \caption{Visualization of the difference in the label assignment during training phase (images are resized to $320\times320$ pixels). Red, yellow and green anchor boxes are positive anchors assigned by $IoU_{anchor}$-based, \emph{Localize to Classify} and \emph{Classify to Localize} respectively. Zoom in to see details.}
\label{fig:visu}
\end{figure}

Here, we would like to explore the reasons for the performance improvements by visualizing the difference in the label assignment between the $IoU_{anchor}$-based strategy and the \emph{Mutual Guidance} strategy during training. Some examples are shown in Figure \ref{fig:visu}. White dotted-line boxes represent ground truth boxes; Red anchor boxes are assigned as positive by $IoU_{anchor}$-based strategy, while considered as negative or ignored by \emph{Localize to Classify} (the top two lines in Figure \ref{fig:visu}) or \emph{Classify to Localize} (the bottom two lines in Figure \ref{fig:visu}); Green anchor boxes are assigned as positive by \emph{Localize to Classify} but negative or ignored by $IoU_{anchor}$-based; Yellow anchor boxes are assigned as positive by \emph{Classify to Localize} but negative or ignored by $IoU_{anchor}$-based.
From these examples, we can conclude that the $IoU_{anchor}$-based strategy only assigns the ``positive'' label to anchors with sufficient IoU with the ground truth box, regardless of their content/context, whereas our proposed \emph{Localize to Classify} and \emph{Localize to Classify} strategies dynamically assign ``positive'' labels to anchors covering semantic discriminant parts of the object (e.g., upper body of a person, main body of animals), and assign ``negative'' labels to anchors with complex background, occluded parts, or anchors containing nearby objects. We believe that our proposed instance-adaptive strategies make the label assignment more reasonable, which is the main reason for performance increase.

\begin{figure}
\begin{center}
\includegraphics[width=120mm]{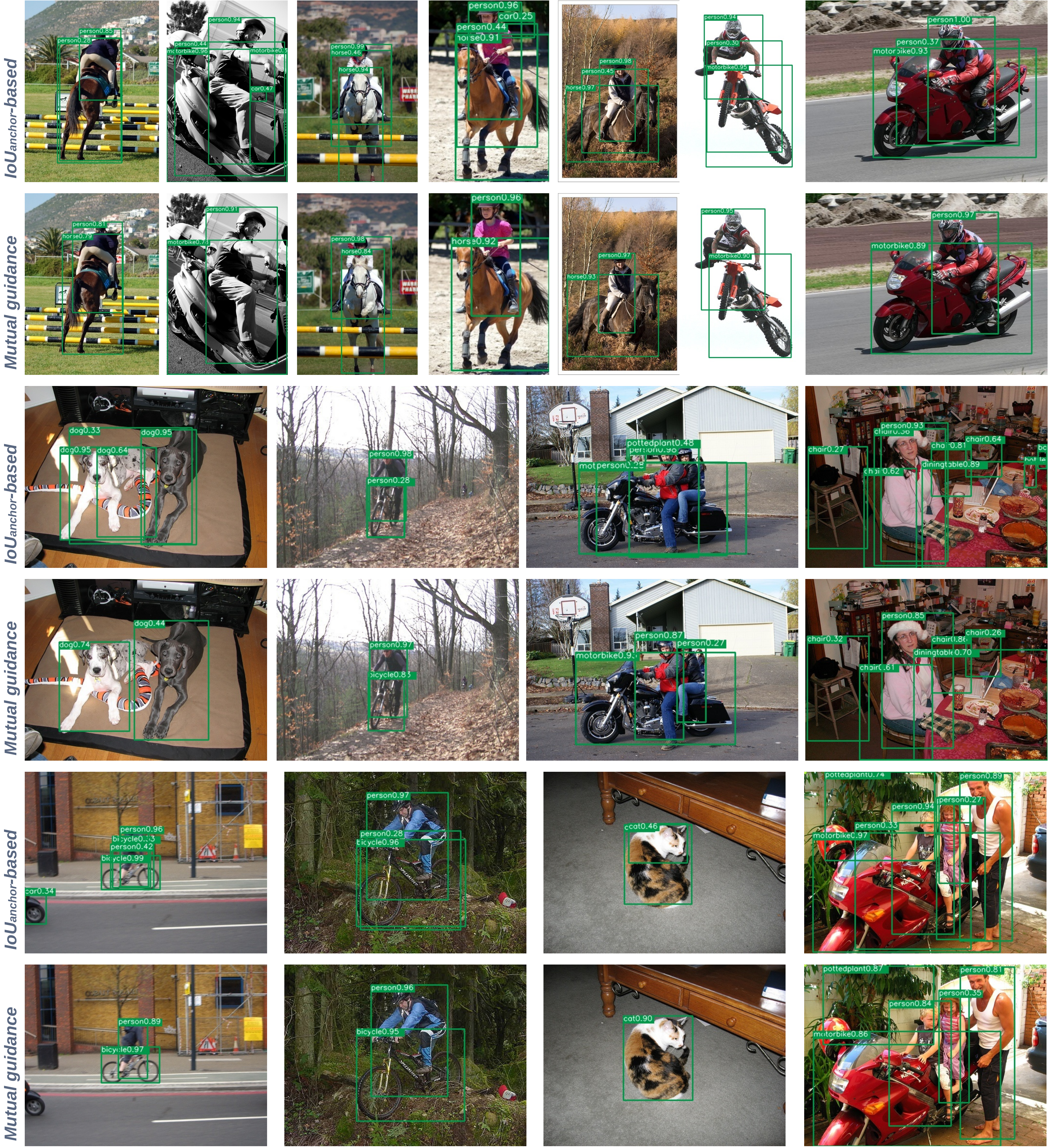}
\end{center}
   \caption{Examples of detection results using an $IoU_{anchor}$-based anchor matching strategy (odd lines) and our proposed \emph{Mutual Guidance} one (even lines). 
   The results are given for all images after applying a Non-Maximum Suppression process with a IoU threshold of $50\%$.
   Zoom in to see details.}
\label{fig:quantitative}
\end{figure}

\paragraph{Detection results visualization.}

Figure \ref{fig:quantitative} illustrates on a few images from the PASCAL VOC dataset the different behaviours shown by our \emph{Mutual Guidance} method and the baseline anchor matching strategy.
As analysed in Section \ref{sec:align}, we can find misaligned predictions (good at classification but poor at localization) from $IoU_{anchor}$-based anchor matching strategy.
As shown in the figure, our method gives better results when different objects are close to each other in the image, e.g. ``man riding a horse'' or ``man riding a bike''.
With the usual $IoU_{anchor}$-based anchor matching strategy, the instance localization and classification tasks are optimized independently of each other. Hence, it is possible that, during the evaluation phase, the classification prediction relies on one object whereas the bounding box regression targets the other object.
However, such a problem is rarer with the \emph{Mutual Guidance} strategy.
Apparently, our anchor matching strategies introduce interactions between both tasks and makes the predictions of localization and classification aligned, which substantially eliminated such false positive predictions.

\section{Conclusion}
\label{sec:conclusion}

In this paper, we question the use of the IoU between predefined anchor boxes and ground truth boxes as a good criterion for anchor matching in object detection and study the interdependence of the two sub-tasks (i.e. localization and classification) involved in the detection process.
We propose a \emph{Mutual Guidance} mechanism, which provides an adaptive matching between anchors and objects by assigning anchor labels for one task according to the prediction quality on the other task and vice versa.
We assess our method on different architectures and different public datasets and compare it with the traditional static anchor matching strategy. Reported results show the effectiveness and generality of this \emph{Mutual Guidance} mechanism in object detection.

\title{Supplementary Material \\ Applying Mutual Guidance to Anchor-free detectors} 
\titlerunning{MutualGuide}
%
\author{Heng ZHANG\inst{1,3}\orcidID{0000-0001-6093-1729} \and
Elisa FROMONT \inst{1,4}\orcidID{0000-0003-0133-3491} \and
Sébastien LEFEVRE\inst{2}\orcidID{0000-0002-2384-8202} \and
Bruno AVIGNON\inst{3}}
\authorrunning{H. ZHANG et al.}
%
\institute{Univ Rennes, IRISA, France \and
Univ Bretagne Sud, IRISA, France \and
ATERMES company, France \and
IUF, Inria, France}
\maketitle

\begin{abstract}
The principle of \emph{Mutual Guidance} in object detection is to assign labels of one task according to the prediction on the other task, and vice versa. Apparently, it is not limited to anchor-based methods, but applicable for any object detector that performs localization and classification tasks. Here we introduce the application of \emph{Mutual Guidance} in anchor-free methods. Experiments conducted on PASCAL VOC dataset demonstrate the consistent precision improvements brought by our method on this category of detectors.
\end{abstract}

\section{Summary of Mutual Guidance}

In order to realize a content/context-sensitive label assignment and avoid the task-misalignment problem, we propose the \emph{Mutual Guidance} mechanism in object detection, which can be summarised as follows:

\begin{itemize}
  \item When assigning labels for the classification task:
  \begin{itemize}
    \item If the prediction of localization is good, the label is assigned as positive (i.e., ``object'') to encourage good predictions on both tasks;
    \item If the prediction of localization is poor, the label is assigned as negative (i.e., ``background'') to avoid misaligned false positive detection;
  \end{itemize}
  \item When assigning labels for the localization task:
  \begin{itemize}
    \item If the prediction of classification is good, the label is assigned as positive (i.e., we optimize this sample) to encourage good predictions on both tasks;
    \item If the prediction of classification is poor, the label is assigned as negative (i.e., we do not optimize this sample) to avoid unnecessary optimizations.
  \end{itemize}
\end{itemize}

\section{Applying Mutual Guidance to FCOS}

Fully Convolutional One-Stage Object Detection (FCOS) \cite{FCOS} is one of the most representative anchor-free methods, which solves object detection in a per-pixel prediction fashion. On each pixel of feature maps, it classifies the category of this sample point and regresses the four distances to the target bounding box borders.
When assigning training labels in FCOS, firstly the corresponding detection layer is selected according to the scale of the object to detect, then positive samples (for both localization and classification tasks) are assigned to all the points inside the ground truth box.
Based on that, \cite{FCOS_PLUS} proposes to only sample the points in the central region of the ground truth box.

When applying \emph{Mutual Guidance} to FCOS, identical to the implementation in anchor-based methods, two strategies are applied: \emph{Localize to Classify} and \emph{Classify to Localize}. Moreover, the same dynamic thresholding strategy is applied. Specifically, for \emph{Localize to Classify}, we firstly note the number of positive samples assigned by the original strategy ($N_p$), then label the $N_p$ highest $IoU_{regressed}$ points as positive; for \emph{Classify to Localize}, we amplify each point's centerness score according to its classification prediction, and label the $N_p$ highest amplified centerness score points as positive. 

Experiments are performed on the PASCAL VOC dataset. ResNet-18 \cite{ResNet} and VGG-16 \cite{VGG} are adopted as backbone networks in our experiments. Unless specified, all other implementation details are the same as in our paper. Experimental results are listed in Table \ref{tab:voc_anchorfree}. Same as with anchor-based methods, our \emph{Mutual Guidance} strategy significantly boosts the detection precision for FCOS, especially on the AP75 metric.

\begin{figure}
\begin{center}
\includegraphics[width=120mm]{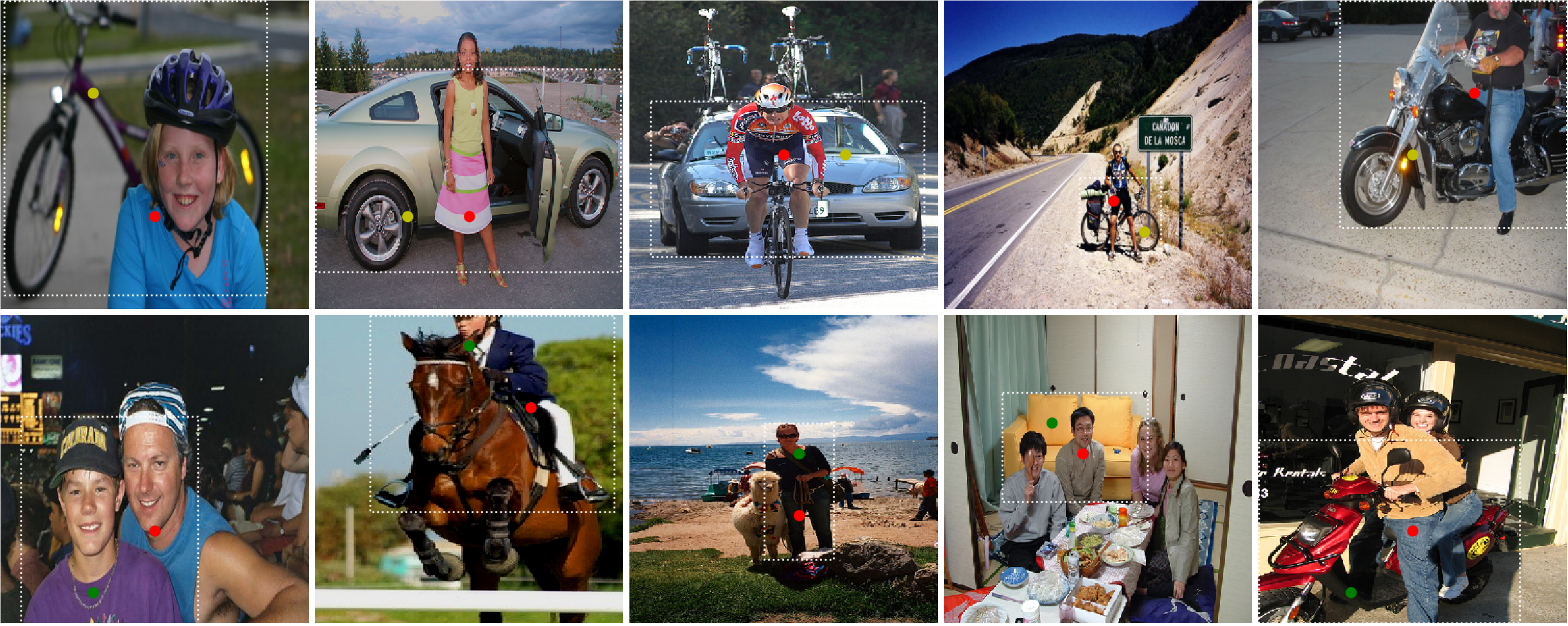}
\end{center}
   \caption{Visualization of the difference in the label assignment during training phase (images are resized to $320\times320$ pixels). The ground truth object in each image is marked by white dotted-line box. Red, yellow and green points are positive samples assigned by original FCOS strategy, \emph{Localize to Classify} and \emph{Classify to Localize} respectively. Zoom in to see details.}
\label{fig:visu_anchorfree}
\end{figure}

\begin{figure}
\begin{center}
\includegraphics[width=120mm]{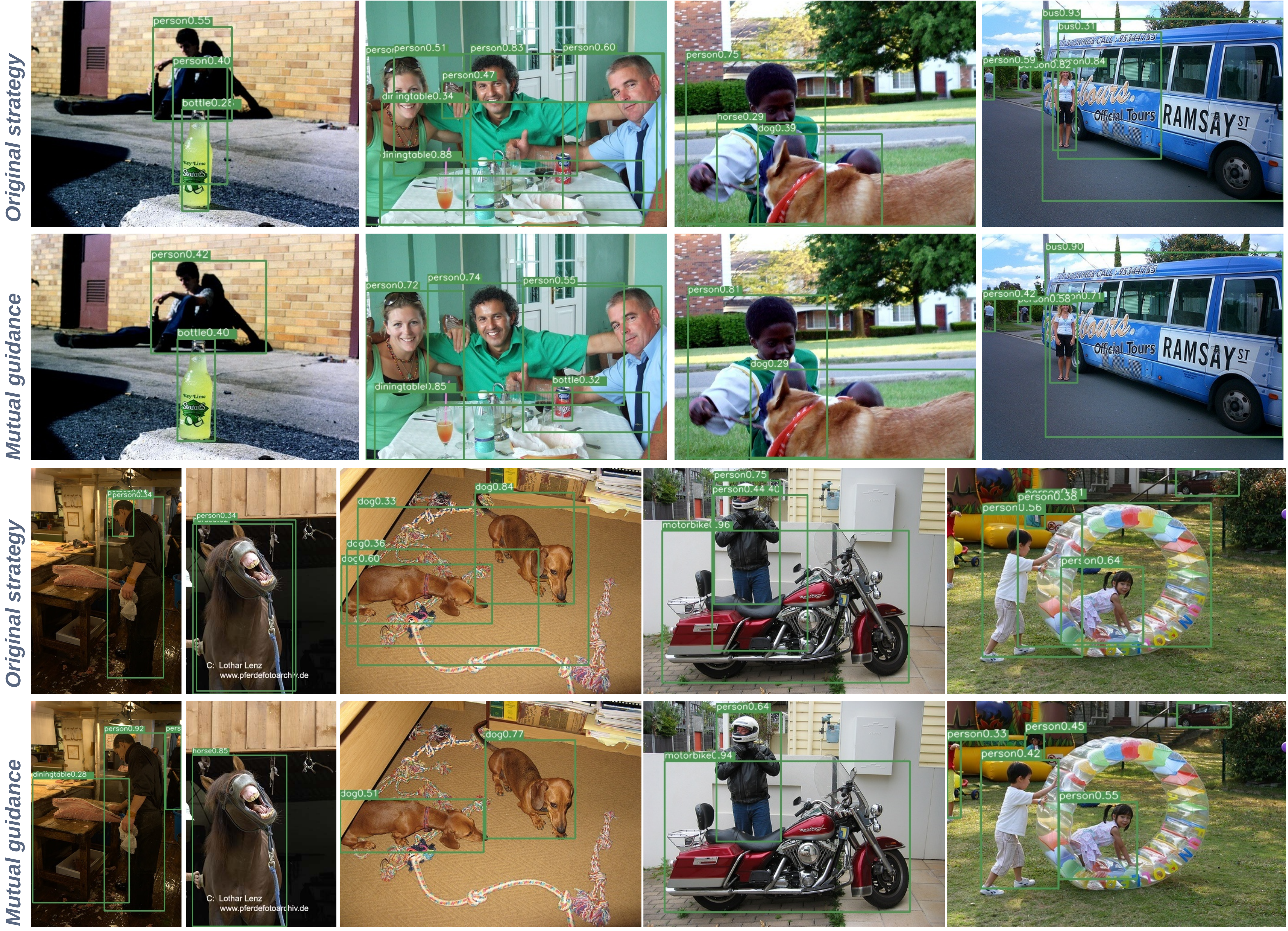}
\end{center}
   \caption{Examples of detection results using the original FCOS label assignment strategy (odd lines) and our proposed \emph{Mutual Guidance} one (even lines). 
   The results are given for all images after applying a Non-Maximum Suppression process with a IoU threshold of $50\%$.
   Zoom in to see details.}
\label{fig:quantitative_anchorfree}
\end{figure}

We then conduct qualitative analysis on the label assignment difference and detection result difference between the original FCOS strategy and \emph{Mutual Guidance}. 
As shown in  Figure \ref{fig:visu_anchorfree}, the original strategy (red points) assigns positive to points in the central region of the object box regardless of their content/context, whereas the \emph{Localize to Classify} (yellow points) and \emph{Classify to Localize} (green points) strategies adaptively assign positive to points with representative semantic information, and assign negative to points on background or nearby objects. 
Since the labels assigned by the original FCOS strategy are always the same for localization and classification tasks, the task-misalignment problem exists in FCOS as well. Several misaligned false positive detections are observed in Figure \ref{fig:quantitative_anchorfree}, however, the proposed \emph{Mutual Guidance} provides more accurate detection.

\begin{table}[t]
\begin{center}
\begin{tabular}{l|c|ccc}
\hline
Model & Matching strategy & $AP$ & $AP50$ & $AP75$ \\ \hline\hline
\multirow{2}{*}{\begin{tabular}[c]{@{}l@{}}FCOS with \\ ResNet-18 backbone\end{tabular}} & original strategy & 49.2\% & 73.7\% & 52.1\%  \\
 & \emph{Mutual Guidance} & \textbf{51.1\%} & \textbf{74.4\%} & \textbf{54.3\%} \\ \hline
\multirow{2}{*}{\begin{tabular}[c]{@{}l@{}}FCOS with \\ VGG-16 backbone\end{tabular}} & original strategy & 53.9\% & 78.4\% & 57.4\% \\
 & \emph{Mutual Guidance} & \textbf{55.9\%} & \textbf{79.4\%} & \textbf{60.2\%} \\ \hline
\end{tabular}
\end{center}
\caption{Comparison of different label assignment strategies (the original one and \emph{Mutual Guidance}) for FCOS. Experiments are conducted on the PASCAL VOC dataset. The best score for each architecture is in bold.}
\label{tab:voc_anchorfree}
\end{table}

\bibliographystyle{splncs}
\bibliography{egbib}

\end{document}